\documentclass[conference]{IEEEtran}
\IEEEoverridecommandlockouts
\usepackage{amsmath,amssymb,amsfonts}
\usepackage[hidelinks]{hyperref}
\usepackage{algorithmic}
\usepackage{graphicx}
\usepackage{textcomp}
\usepackage[linesnumbered,ruled,vlined]{algorithm2e}
\usepackage{balance}
\usepackage{authblk}
\usepackage[table]{xcolor}
\usepackage{booktabs}
\usepackage{xcolor}
\usepackage{orcidlink}
\usepackage[export]{adjustbox}
\usepackage{cite}
\usepackage{orcidlink}
\usepackage{amsmath,amssymb,amsfonts}
\usepackage{algorithmic}
\usepackage{graphicx}
\usepackage{subcaption}
\usepackage{textcomp}
\usepackage{xcolor}
\usepackage{pifont}

\newcommand{\blue}[1]{#1}

\def\BibTeX{{\rm B\kern-.05em{\sc i\kern-.025em b}\kern-.08em
    T\kern-.1667em\lower.7ex\hbox{E}\kern-.125emX}}
\begin{document}

\title{PANDORA: Pixel-wise Attention Dissolution and Latent Guidance for Zero-Shot Object Removal}

\author[1,2]{Dinh-Khoi Vo \orcidlink{0000-0001-8831-8846}}
\author[1,2]{Van-Loc Nguyen \orcidlink{0000-0001-9351-3750}}
\author[3]{Tam V. Nguyen \orcidlink{0000-0003-0236-7992}}
\author[1,2]{Minh-Triet Tran \orcidlink{0000-0003-3046-3041}}
\author[1,2]{Trung-Nghia Le \orcidlink{0000-0002-7363-2610}$^\dagger$\thanks{$^\dagger$Coressponding author: ltnghia@fit.hcmus.edu.vn}}

\affil[1]{University of Science, Ho Chi Minh City, Vietnam}
\affil[2]{Vietnam National University, Ho Chi Minh City, Vietnam}
\affil[3]{University of Dayton, Ohio, United States}

\maketitle

\begin{abstract}
Removing objects from natural images is challenging due to difficulty of synthesizing semantically coherent content while preserving background integrity. Existing methods often rely on fine-tuning, prompt engineering, or inference-time optimization, yet still suffer from texture inconsistency, rigid artifacts, weak foreground–background disentanglement, and poor scalability for multi-object removal. We propose a novel zero-shot object removal framework, namely PANDORA, that operates directly on pre-trained text-to-image diffusion models, requiring no fine-tuning, prompts, or optimization. We propose Pixel-wise Attention Dissolution to remove object by nullifying the most correlated attention keys for masked pixels, effectively eliminating the object from self-attention flow and allowing background context to dominate reconstruction. We further introduce Localized Attentional Disentanglement Guidance to steer denoising toward latent manifolds favorable to clean object removal. Together, these components enable precise, non-rigid, prompt-free, and scalable multi-object erasure in a single pass. Experiments demonstrate superior visual fidelity and semantic plausibility compared to state-of-the-art methods. The project page is available at \url{https://vdkhoi20.github.io/PANDORA}.
\end{abstract}


\section{Introduction}

Recent advances in diffusion models~\cite{dhariwal2021diffusion,Hoang2025ShowFlow} have revolutionized generative image modeling, enabling high-quality image synthesis guided by textual prompts or other conditions. These models have opened new opportunities for semantic image editing~\cite{hertz2022prompt,parmar2023zeroshot}, particularly in longstanding and challenging object removal ~\cite{suvorov2022resolution,sun2025attentive,ekin2024clipaway} that involves selectively erasing undesired regions from an image and reconstructing the void with content that blends naturally into the scene. This task goes beyond typical inpainting by requiring both precise object elimination and faithful restoration that aligns with the scene’s semantics and visual flow.

Conventional object removal techniques have mainly utilized patch-based approaches~\cite{zomet2003learning,barnes2009patchmatch} or GANs~\cite{suvorov2022resolution,li2022mat}. Patch-based methods often result in inconsistencies by filling in the masked areas with patches from other parts of the image, which can lead to unnatural blending with the surrounding regions. GANs, while enhancing realism, still face challenges with artifact generation and lack versatility in handling complex scenes. 

\begin{figure}[!t]
    \centering
    \includegraphics[width=\linewidth]{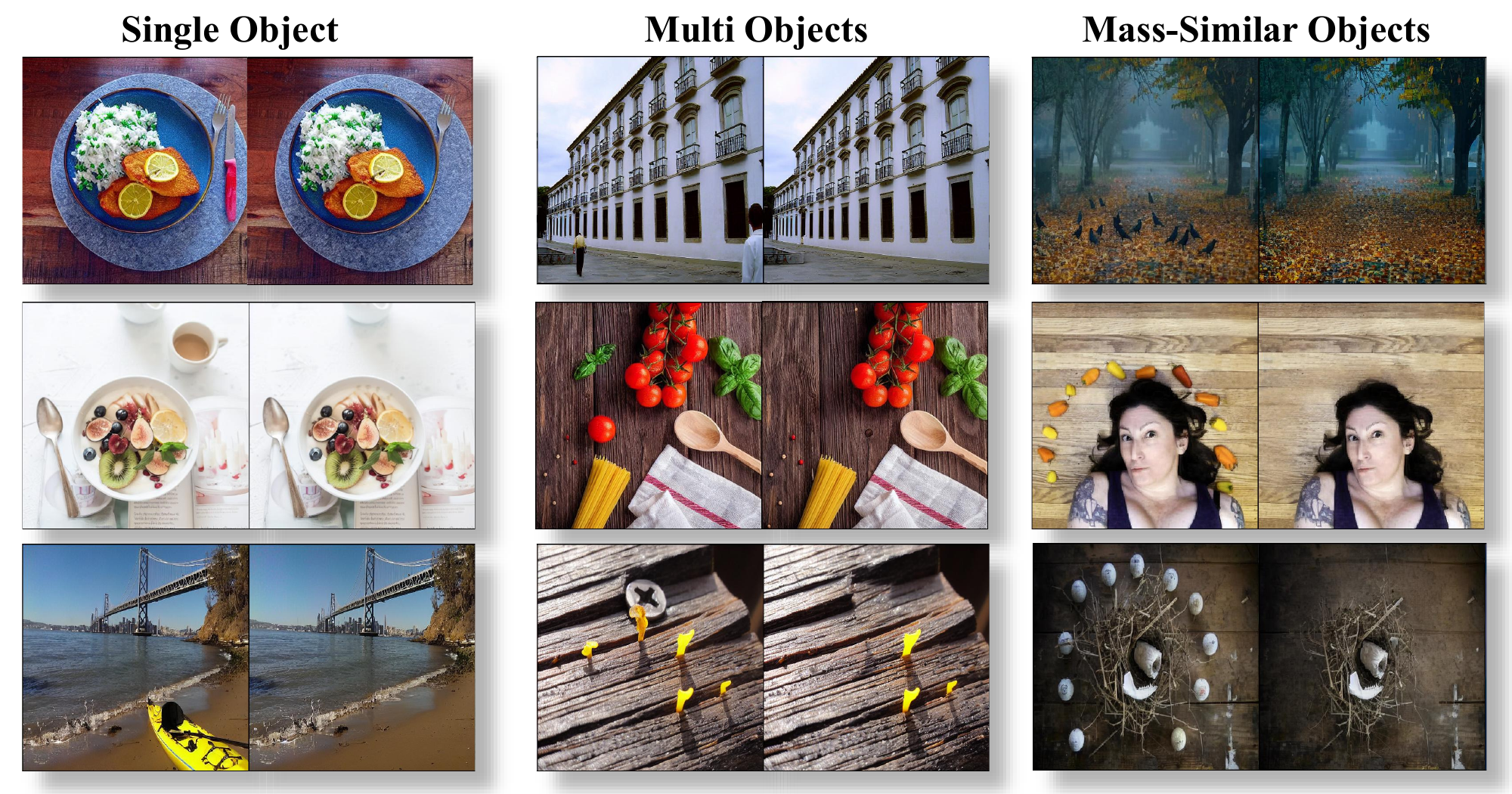}
    \caption{Our PANDORA enables prompt-free, fine-tuning-free object removal across various scenarios in a single forward. Without requiring training or textual prompts, our approach handles diverse and challenging removal settings, from a single object to multiple similar or distinct targets and even densely packed similar objects, while preserving background fidelity and structural consistency.
    }
    \label{fig:teaser}
    \vspace{-5mm}
\end{figure}

Recent methods based on diffusion models~\cite{Nguyen2026Neurocomputing} have significantly improved the realism of image synthesis. However, when adapted to object removal, these models often struggle to reliably eliminate target objects. One such adaptation, Stable Diffusion Inpainting~\cite{rombach2022highresolution}, extends the base model by conditioning on a binary mask to enable end-to-end inpainting. Despite its tailored design, the model frequently produces incomplete removals or introduces unexpected artifacts, even with considerable fine-tuning effort. Moreover, they require extensive prompt engineering~\cite{brooks2022instructpix2pix} or fully fine-tuning such large-scale models~\cite{rombach2022highresolution} is often impractical in low-resource settings, limiting their accessibility and scalability in broader research applications. On the other hand, zero-shot methods~\cite{sun2025attentive,vo2025cpam} have been proposed but they often struggle to handle multi-object removal, leading to incomplete removals or noticeable artifacts. CPAM~\cite{vo2025cpam} can preserve background context but often fail to fully erase objects due to residual attention. Attentive Eraser~\cite{sun2025attentive} attempts to scale down attention weights but uses a uniform strategy that ignores pixel-wise correlations, often requiring sensitive manual tuning.



To address these challenges, we propose PANDORA, a novel zero-shot object removal framework that operates directly on pre-trained diffusion models in a single pass, without any fine-tuning, prompt engineering, or inference-time optimization, thus fully leveraging their latent generative capacity for inpainting. The input image is inverted into the latent noise space via a preservation adaptation module~\cite{vo2025cpam} to ensure that the background regions remain unaffected during the object removal process, allowing targeted and artifact-free editing. We also introduce two synergistic core modules: Pixel-wise Attention Dissolution (PAD) dissolves object information at the attention level and Localized Attentional Disentanglement Guidance (LADG) complements it by reshaping the denoising trajectory in latent space, ensuring that masked regions are reconstructed coherently with their surrounding context. Particularly, PAD enables fine-grained control of self-attention by removing the top-k most correlated keys via percentile-based thresholding. This disconnects masked query pixels from dominant foreground regions, dissolving object information and enabling precise, natural reconstruction with coherent background content. Meanwhile, LADG steers latent noise away from object regions while preserving surrounding areas, refining the denoising trajectory in latent space to suppress residual artifacts and produce smoother, cleaner results.


To evaluate multiple object removal methods, we build a new benchmark dataset, comprising 75 single-object samples, 17 multi-object samples, and 94 mass-similar object samples, with masks obtained through manual annotation or automatic extraction. Extensive experiments show that PANDORA excels in complex and cluttered scenes, supports multi-object erasure, and requires no prompts or training-time supervision. 

Our contributions are summarized as follows:
\begin{itemize}
    \item We propose PANDORA, a novel zero-shot object removal framework that directly leverages pre-trained diffusion models to remove objects in a single pass without fine-tuning, prompt engineering, or inference-time optimization.
    
    \item We propose Pixel-wise Attention Dissolution (PAD) to dissolve object information by disconnecting masked query pixels from their most correlated keys in self-attention, enabling fine-grained removal and reconstruction with coherent background content.
    
    \item We present Localized Attentional Disentanglement Guidance (LADG), which steers latent noise away from object regions while preserving unaffected areas, refining the denoising trajectory to remove residual artifacts and produce cleaner outputs.
    
    \item We build a challenging benchmark dataset for evaluating object removal methods. Experiments demonstrate that our method achieves SOTA zero-shot object removal performance, excelling in both single- and multi-object scenarios.
\end{itemize}

\begin{figure*}[!t]
    \centering
    \includegraphics[width=0.7\textwidth]{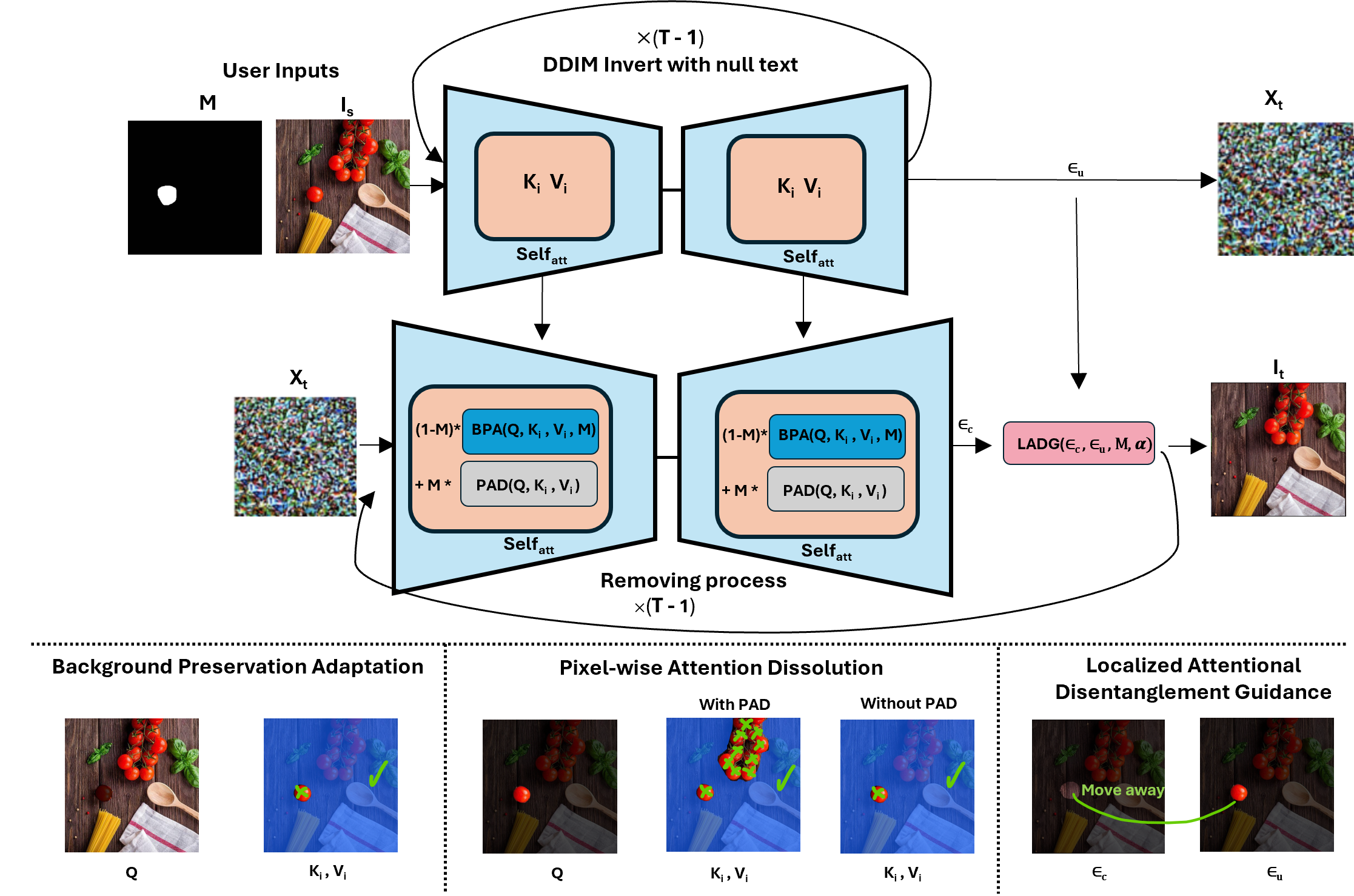}
    \caption{Overview of our proposed pipeline with an intuitive illustration of each module. The image is inverted into noise with intermediate latents stored and injected into BPA and PAD to preserve background and dissolve objects, respectively. Specifically, BPA restricts background queries to background regions, while PAD operates at the pixel level to constrain object queries to unrelated regions. Finally, LADG steers denoising away from masked object regions for seamless synthesis.}
    \label{fig:overview}
    \vspace{-3mm}
\end{figure*}

\begin{algorithm}[!t]
\small
\caption{PANDORA's algorithm: Zero-Shot Object Removal}\label{algo:overview}
\textbf{Inputs:} 
A mask \(M\), the intermediate latent noises \(x^i\), the target initial latent noise map \(x_T\). \\
\textbf{Output:} Erased Image \(I_t\).

\begin{enumerate}
    \item \textbf{For} \(t = T, T-1, \ldots, 1\) \textbf{do:}
    \begin{enumerate}
        
        \item \(\{\_,K_i, V_i\} \overset{\text{get}}{\longleftarrow} \epsilon_\theta(x^i_t, t)\)
        \item \(\{Q, \_, \_\} \overset{\text{get}}{\longleftarrow} \epsilon_\theta(x_t, t)\)
        \item \(\text{self-attention} \overset{\text{inject}}{\longleftarrow} \text{BPA}(\text{ \(Q, K_i, V_i, M\)})\)
        \item \(\text{self-attention} \overset{\text{inject}}{\longleftarrow} \text{PAD}(\text{ \(Q, K_i, V_i, M, top_k\)})\)
        
        \item \(\epsilon \leftarrow \epsilon_\theta(x_t, t,\text{self-attention})\)
        \item \(\epsilon \leftarrow \text{LADG}( \epsilon, \epsilon_\theta(x^i_t, t), M, \alpha_t )\)
        \item \(x_{t-1} \leftarrow \text{Sample}(x_t, \epsilon)\)
      
    \end{enumerate}
    \item \textbf{End For.}
\end{enumerate}
\textbf{Return:} VAE(\(x_0\)).

\end{algorithm}

\section{Proposed Method}
\subsection{Overview}
PANDORA performs zero-shot object removal directly on a pre-trained diffusion model. Given an input image $I_s$ and a binary mask $M$ specifying the target objects, the model produces an edited image $I_t$ where the masked regions are erased and seamlessly reconstructed with contextually consistent background. The process begins with latent inversion to map the input image into the noise space while preserving unaffected regions in the denoising process. We then apply Pixel-wise Attention Dissolution (PAD) to disconnect masked query pixels from their most correlated keys, effectively dissolving object information at the attention level. Next, Localized Attentional Disentanglement Guidance (LADG) steers the denoising trajectory in latent space away from the object regions, refining the reconstruction to suppress residual artifacts. 
Together, PAD and LADG enable precise, pixel-level control for single- and multi-object removal in a single forward pass, without any fine-tuning, prompt engineering, or inference-time optimization. Figure~\ref{fig:overview} illustrates the overall pipeline and algorithm is outlined in Algorithm~\ref{algo:overview}.

\subsection{Background Preservation Adaptation (BPA)}
To ensure that non-masked regions remain unaffected during the object removal process, we adopt the preservation adaptation module from CPAM~\cite{vo2025cpam} for background consistency. Specifically, we first invert the input image $I_s$ into the latent noise space using DDIM inversion~\cite{song2020denoising}, which reconstructs the noise trajectory deterministically. During this process, the intermediate latent states $x_i$ are stored at each timestep $t$ to preserve semantic information.  

At a denoising step $t$, let $(Q, K, V)$ denote the query, key, and value features of the current noise in Unet's self-attention, to retain the background unchanged, we get $(K_i, V_i)$ the key and value features extracted from the stored latent noise at step $t$ while retaining the query feature $Q$. The background semantic content $SC_{bg}$ is obtained by applying mask-guided attention:  

\begin{equation}
SC_{bg} = \text{Att}(Q, K_i, V_i; 1 - M),
\end{equation}
where $\text{Att}(\cdot)$ denotes the attention mechanism and $(1 - M)$ ensures that only non-masked (background) areas contribute to the attention computation.  

This operation effectively transfers semantic content from the original latent noise to the background region of the edited image, thereby preserving structural and visual fidelity in areas outside the object mask. As a result, the subsequent object removal modules can focus exclusively on masked regions without degrading the integrity of the surrounding scene.

\subsection{Pixel-wise Attention Dissolution (PAD)}
\label{subsec:pad}

In contrast to background preservation, which reuses $(K_i, V_i)$ to retain non-masked content, PAD aims to erase object regions specified by the mask $M$ by dissolving their strongest semantic connections in self-attention. Given the query features $Q$ and stored $(K_i, V_i)$, the attention logits are first computed as:  

\begin{equation}
\mathbf{S}_{t,l} = \frac{Q K_i^\top}{\sqrt{d}},
\end{equation}
where $d$ is the feature dimension, and $t, l$ denote the denoising step and self-attention layer in the U-Net.
To suppress dominant associations, we apply percentile-based thresholding over each query’s attention distribution and set the top-$k$ strongest connections to $-\infty$:  

\begin{equation}
\mathbf{S}^{\text{diss}}_{t,l}[i,j] =
\begin{cases}
-\infty & \text{if } j \in \text{Top-}k(\mathbf{A}_{t,l}[i,:]) \ \lor \ j \in M, \\
\mathbf{S}_{t,l}[i,j] & \text{otherwise},
\end{cases}
\end{equation}
where $\mathbf{A}_{t,l} = \text{softmax}(\mathbf{S}_{t,l})$ is the normalized attention map.  
The dissolved self-attention output is then computed as:  

\begin{equation}
\mathbf{B}_{t,l} = \text{softmax}(\mathbf{S}^{\text{diss}}_{t,l}) \cdot V_i.
\end{equation}

Finally, to integrate PAD with background preservation, the final output representation is obtained by blending object and background content:  

\begin{equation}
\mathbf{O}_{t,l} = \mathbf{B}_{t,l} \odot M + \mathbf{SC}^{\text{bg}}_{t,l} \odot (1 - M),
\end{equation}
where $\mathbf{SC}^{\text{bg}}_{t,l}$ denotes the background-preserved semantic content.  

Through this pixel-wise dissolution and selective recombination, each masked query abandons its strongest semantic ties, allowing the object region to vanish and be naturally reconstructed from surrounding background context, while ensuring global scene consistency.

\subsection{Localized Attentional Disentanglement Guidance (LADG)}
\label{subsec:ladg}

Diffusion models are score-based generators: their denoising trajectory is guided by noise predictions that approximate the gradient of the data distribution. Existing guidance strategies such as classifier-based~\cite{dhariwal2021diffusion} (requiring an auxiliary classifier) or classifier-free~\cite{ho2021classifierfree} (requiring training with conditional vs. unconditional embedding) operate globally. In contrast, our approach requires no auxiliary classifier or re-training with conditional and unconditional embedding; instead, we introduce LADG, a spatially gated mechanism that \emph{pushes the latent distribution away from the original (unconditioned) trajectory only inside the object mask}. Denote the unconditional and conditional noise predictions at step $t$ by:
\[
\epsilon_u(x^i_t,t):=\epsilon_\theta(x^i_t,t;\varnothing),\qquad
\epsilon_c(x_t,t):=\epsilon_\theta(PAD(x_t),t;\varnothing)
\]
Rather than applying a global guidance scale, LADG forms a \emph{mask-aware} noise prediction:

\begin{equation}
\begin{aligned}
\hat{\epsilon}(x_t,t)
&= (1-M)\odot \epsilon_c(x_t,t) \\
&\quad + M\odot\Big[\alpha_t\,\epsilon_c(x_t,t) 
       + (1-\alpha_t)\,\epsilon_u(x^i_t,t)\Big],
\end{aligned}
\label{eq:ladg_masked}
\end{equation}
where $\alpha_t$ is a real number controlling the degree to which masked latents are steered away from the unconditional trajectory and toward the conditional prediction.

Intuitively, outside the mask we preserve the conditional trajectory to maintain scene fidelity, while inside the mask we explicitly drive the latents away from the original object-unconditioned score, thereby accelerating object removal and mitigating residual artifacts to produce cleaner and more coherent outputs.

\section{Experiments}
\subsection{Implementation Details}
PANDORA is model-agnostic and can be applied to various diffusion backbones. In this work, we built it upon Stable Diffusion v1.5 and v2.1 using the official weights from Hugging Face diffusers. We adopted DDIM inversion with 50 denoising steps. The proposed PAD module was integrated into the U-Net’s self-attention layers without any retraining. Specifically, PAD suppressed the top 2–5\% of correlated activations within masked regions to decouple object dependencies, while LADG applied a latent guidance weight $\alpha_t$ typically in the range $[1.0,\,1.6]$ to stabilize background refinement. To avoid disconnection artifacts, BPA and PAD were only activated during the early denoising phase (steps 1–40/45), after which full self-attention was restored for final refinement.

\subsection{Benchmark Dataset}
To evaluate methods, we constructed a new benchmark dataset for multi-object removal by collecting images from multiple sources. Specifically, single- and multi-object samples were taken from PIE-Bench~\cite{ju2023direct} and Open Images Dataset~\cite{kuznetsova2020open}, both of which provide high-quality paired images and corresponding object masks. For the mass-similar object removal scenario, we included samples from Ranjan et al.~\cite{ranjan2021learning}, which feature clusters of visually similar objects such as fruits, flowers, and birds. We removed low-quality or overly simple samples and retained diverse scenes with varied object shapes, textures, and contextual relationships. In total, the dataset comprises 75 single-object samples, 17 multi-object samples, and 94 mass-similar object samples, with masks obtained through manual annotation or automatic extraction.

\subsection{Evaluation Metrics}
To comprehensively evaluate object removal performance, we assess both background fidelity and object removal quality.
For background regions, we employed Mean Squared Error (MSE) and LPIPS~\cite{zhang2018unreasonable} to quantify the pixel-level and perceptual similarity between the generated background and the ground truth background.
To measure how effectively the target object is removed, we adopted the CLIP score~\cite{clip}, which evaluates the semantic alignment between the edited region and a background-related text prompt (i.e., “A background without any objects.”). In addition, we computed the Fréchet Inception Distance (FID)~\cite{heusel2017gans} to assess the overall realism and distributional consistency of the generated images compared to the ground truth background regions (excluding masked areas). To ensure a fair comparison across all methods, all images were resized to 512×512 resolution, and the masked regions were uniformly processed for each dataset.

\subsection{Compared Methods}
We compared PANDORA with a diverse set of mask-guided state-of-the-art object removal methods, covering both fine-tuned and zero-shot paradigms.
For fine-tuned methods, we included LaMa~\cite{suvorov2022resolution}, PowerPaint~\cite{zhuang2024task}, and SD2-Inpaint~\cite{rombach2022highresolution}, evaluated with and without text prompts. Specifically, PowerPaint v2.1 is guided by the prompt “empty scene blur”, while SD2-Inpaint uses “a clean, natural background, seamless and realistic lighting.”
For zero-shot methods, we evaluated CPAM~\cite{vo2025cpam}, Attentive Eraser~\cite{sun2025attentive}, and our proposed PANDORA. All are built upon Stable Diffusion v1.5 and v2.1, except Attentive Eraser, whose officially public implementation does not support the v2.1 checkpoint.
For all methods based on the Stable Diffusion backbone (except LAMA), we set 50 denoising steps for consistency.
All methods are tested under identical masking conditions for fair comparison, using their official implementations and default configurations.

\begin{figure*}[!t]
    \centering
    \includegraphics[width=0.9\textwidth]{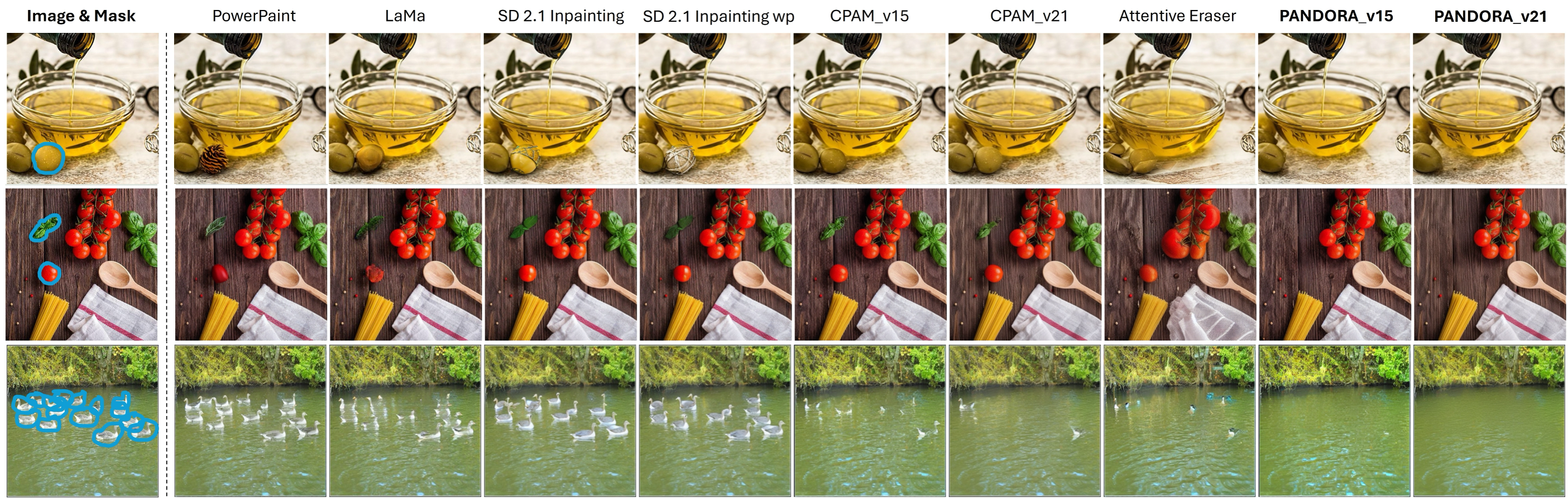}
    \caption{Qualitative comparison across diverse object removal scenarios, including single-object, multi-object, and mass similar-object removal (top to bottom). From left to right: the original image with mask and results from different methods. The last five columns correspond to zero-shot approaches.}
    \label{fig:qualitative}
    \vspace{-3mm}
\end{figure*}

\subsection{Qualitative Results}

Figure~\ref{fig:qualitative} presents qualitative comparisons across various scenarios. PANDORA effectively removes both distinct and clustered objects while preserving fine background details and maintaining global coherence. In contrast, fine-tuning-based methods often modify or replace target objects instead of removing them. Among zero-shot methods, CPAM~\cite{vo2025cpam} preserves background structure relatively well, whereas Attentive Eraser~\cite{sun2025attentive} tends to lose fine details or generate overly altered textures; however, both struggle to erase objects when similar instances appear elsewhere in the scene. All methods except PANDORA frequently introduce residual artifacts or distortions, while PANDORA consistently produces clean, contextually coherent reconstructions across both Stable Diffusion v1.5 and v2.1 backbones, demonstrating its strong generalization and stability.

\subsection{Quantitative Results}

\begin{table}[!t]
\centering
\caption{Quantitative comparison of fine-tuned and zero-shot methods. PANDORA consistently achieves the best object removal quality with competitive background realism, without any retraining or textual prompts, demonstrating strong generalization across bdifferent backbones. Removing LADG slightly reduces removal quality, while removing PAD causes a significant degradation.}
\label{tab:quantitative_comparison}
\resizebox{\linewidth}{!}{
\begin{tabular}{lccccc}
\toprule
\textbf{Method}  & \textbf{Text} & \textbf{FID}↓ & \textbf{LPIPS}↓ & \textbf{MSE}↓ & \textbf{CLIP score}↑ \\
\midrule
\multicolumn{6}{c}{\textit{Fine-tuning-based methods (SD 2.1 backbone, except LaMa)}} \\
\midrule
PowerPaint~\cite{zhuang2024task}  & \ding{51} & 22.81 & 0.1322 & 0.0104 & 24.15 \\
LaMa~\cite{suvorov2022resolution}  & \ding{55} & \textbf{0.71} & \textbf{0.0012} & \textbf{0.0001} & 24.50 \\
SD2-Inpaint~\cite{rombach2022highresolution} & \ding{55} & 17.93 & 0.1106 & 0.0073 & 24.06 \\
SD2-Inpaint-w/ prompt~\cite{rombach2022highresolution} & \ding{51} & 18.01 & 0.1098 & 0.0072 & 24.32 \\
\midrule[0.3pt]
\multicolumn{6}{c}{\textit{Zero-shot methods (no retraining, SD 2.1 backbone)}} \\
\midrule
CPAM~\cite{vo2025cpam}  & \ding{55} & 25.25 & 0.0953 & 0.0048 & 24.49 \\
\textbf{PANDORA w/o PAD}  & \ding{55} & 27.3 & 0.0985 & 0.005 & 24.58 \\
\textbf{PANDORA w/o LADG}  & \ding{55} & 30.8 & 0.1007 & 0.0055 & 24.65 \\
\textbf{PANDORA}  & \ding{55} & 35.1 & 0.1064 & 0.0059 & \textbf{24.69} \\

\midrule[0.3pt]
\multicolumn{6}{c}{\textit{Zero-shot methods (no retraining, SD 1.5 backbone)}} \\
\midrule
CPAM~\cite{vo2025cpam}  & \ding{55} & 29.54 & 0.1564 & 0.0138 & 24.32 \\
Attentive Eraser~\cite{sun2025attentive} & \ding{55} & 118.09 & 0.2567 & 0.0270 & 24.42 \\
\textbf{PANDORA w/o PAD}  & \ding{55} & 35.59 & 0.1702 & 0.0156 & 24.4 \\
\textbf{PANDORA w/o LADG}  & \ding{55} & 42.17 & 0.1844 & 0.0171 & 24.55 \\
\textbf{PANDORA}  & \ding{55} & 44.98 & 0.1895 & 0.0184 & 24.57 \\

\bottomrule
\end{tabular}
}
\vspace{-5mm}
\end{table}

Table~\ref{tab:quantitative_comparison} presents the averaged quantitative results across all dataset types. Among zero-shot methods, PANDORA achieves the best balance between background realism and object removal quality across both Stable Diffusion v1.5 and v2.1 backbones. It attains the highest CLIP score, reflecting effective object elimination, while maintaining competitive LPIPS and MSE values. CPAM~\cite{vo2025cpam} delivers balanced background reconstruction but struggles to fully erase objects, whereas Attentive Eraser~\cite{sun2025attentive} often distorts scene structures, leading to high FID, LPIPS, and MSE values. For fine-tuned models, LaMa~\cite{suvorov2022resolution} achieves the lowest FID, LPIPS, and MSE scores due to its rigid blending strategy that merges original background patches, limiting realistic restoration. PowerPaint~\cite{zhuang2024task} and SD2-Inpaint~\cite{rombach2022highresolution} yield moderate performance but depend on task-specific training or textual prompts. Overall, PANDORA demonstrates consistent performance across different diffusion backbones, achieving competitive background coherence and the most effective object removals.

\begin{table}[!t]
\centering
\caption{User study results showing the User Preference Rate (UPR) as the percentage of times each method was chosen as the best. PANDORA is consistently favored by users.}
\label{tab:userstudy}
\begin{tabular}{lr}
\toprule
\textbf{Method} & \textbf{UPR} \\
\midrule
\multicolumn{2}{c}{\textit{Fine-tuning-based methods}} \\
\midrule
PowerPaint~\cite{zhuang2024task} & 4.50 \\
LaMa~\cite{suvorov2022resolution} & 6.75 \\
SD2-Inpaint~\cite{rombach2022highresolution} & 6.25 \\
SD2-Inpaint-w/ prompt~\cite{rombach2022highresolution} & 7.00 \\
\midrule[0.3pt]
\multicolumn{2}{c}{\textit{Zero-shot methods (no training required)}} \\
\midrule
CPAM~\cite{vo2025cpam} & 6.00 \\
Attentive Eraser~\cite{sun2025attentive} & 22.00 \\
\textbf{PANDORA (Ours)} & \textbf{47.50} \\
\bottomrule
\end{tabular}
\vspace{-5mm}
\end{table}

Ablation results in Table~\ref{tab:quantitative_comparison} show that removing LADG slightly reduces removal accuracy, as the absence of localized guidance weakens contextual refinement. In contrast, omitting PAD leads to a significant performance drop, as correlated activations within masked regions are not effectively suppressed, often resulting in incomplete erasure or even failure to remove the target object.

\subsection{User Study}

To evaluate the effectiveness of PANDORA, we conducted a user study with 20 participants from diverse backgrounds. Each participant was asked to select the best image from sets of outputs, where the original image and the results of six different methods were presented side-by-side. To ensure objectivity, the methods were shuffled and blinded so participants did not know which image corresponded to which method, including PANDORA. The evaluation was organized into 20 batches, each containing 20 randomly selected samples, yielding a total of 400 responses and 2,800 image considerations across all methods. Then, we report the User Preference Rate (UPR), defined as the percentage of participants selecting a given method.

Table~\ref{tab:userstudy} summarizes results, showing that PANDORA was consistently favored by users (with the highest UPR). The user study reinforces our qualitative and quantitative findings, highlighting PANDORA’s effectiveness in both single-and multi- object removal.

\begin{figure}[!t] 
\centering
\includegraphics[width=\linewidth]{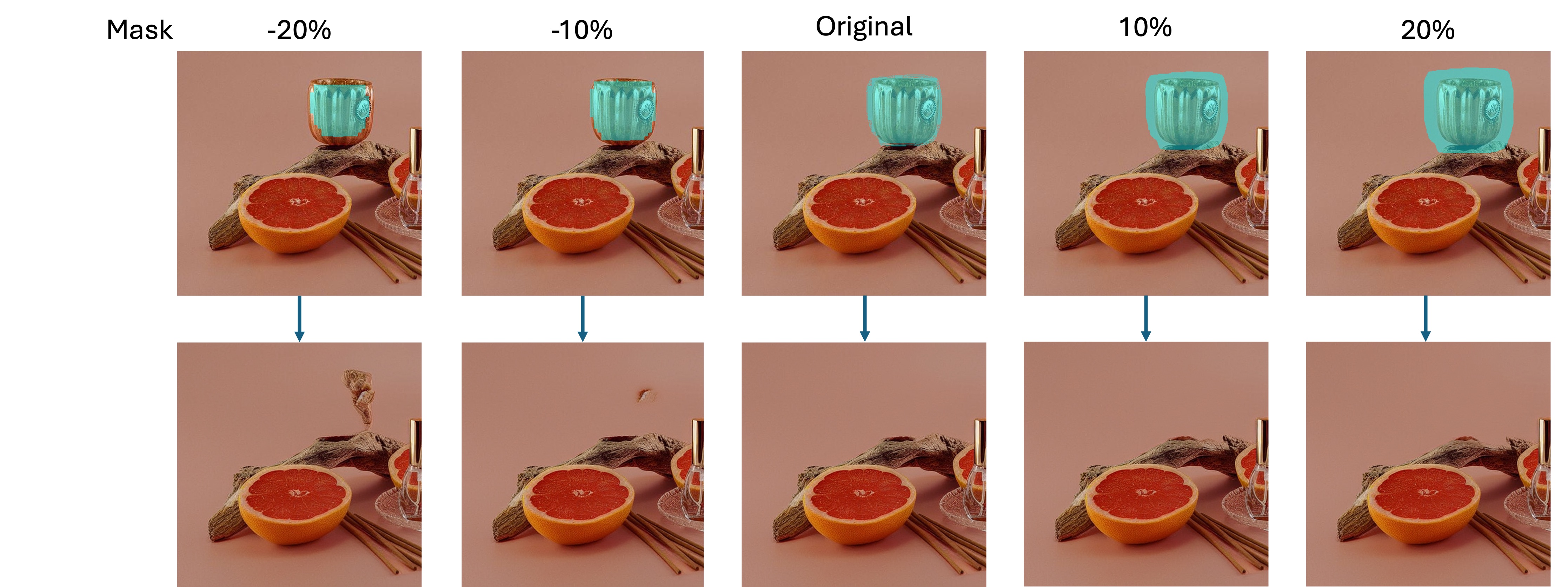}
    \caption{\blue{From left to right, mask is $20\%$ smaller, $10\%$ smaller, tightly aligned, $10\%$ larger, and $20\%$ larger than object.}} \label{fig:mask}
\end{figure}

\subsection{Limitations \blue{and Discussion}}
Although effective, our approach has several limitations. First, suppression based on a fixed percentile threshold may occasionally over-filter or under-filter attention responses, leading to incomplete or excessive object removal. \blue{Nevertheless, our empirical results indicate that suppressing the top 2–5\% of correlations achieves a favorable balance between removal quality and visual fidelity, providing the most stable overall performance (see Tab.~\ref{tab:PAD})}. Second, the framework relies on accurate binary masks; imprecise segmentation can degrade disentanglement quality and introduce artifacts. \blue{Undersized masks may leave residual artifacts, while moderately oversized masks are generally well tolerated (see Fig.~\ref{fig:mask})}. In future work, we plan to explore adaptive mechanisms that adjust percentile thresholds according to attention statistics or mask confidence, as well as automatic region selection to identify removable objects without manual input.

\section{Conclusion}
In this paper, we introduced PANDORA, a novel zero-shot framework for object removal that operates directly on pre-trained diffusion models without requiring any fine-tuning, prompts, or inference-time optimization. Our approach addresses the key challenges of maintaining background integrity while achieving clean, semantically coherent object erasure. The core of our method lies in two synergistic components: Pixel-wise Attention Dissolution (PAD), which precisely dissolves object information at a granular level by nullifying the most correlated keys in self-attention, and Localized Attentional Disentanglement Guidance (LADG), which steers the denoising process away from object-related latent manifolds. Together, these modules enable flexible and effective removal of single, multiple, and even densely packed objects in a single forward pass. Extensive experiments demonstrate that PANDORA significantly outperforms existing state-of-the-art methods, including those that rely on fine-tuning and prompt guidance, setting a new benchmark for zero-shot object removal in terms of both visual quality and semantic plausibility.

\begin{table}[!t]
\centering
\caption{\blue{Ablation study on percentile threshold used in PAD (SD 1.5 backbone).}}
\label{tab:PAD}
\resizebox{0.48\textwidth}{!}{
\begin{tabular}{c c c c c}
\toprule
\textbf{Suppressed Percentile} & \textbf{FID}↓ & \textbf{LPIPS}↓ & \textbf{MSE}↓ & \textbf{CLIP}↑ \\
\midrule
1\%   & \textbf{42.32} & \textbf{0.1837} & \textbf{0.0175} & 24.547 \\
3\%   & \textit{44.07} & \textit{0.1871} & \textit{0.0180} & 24.553 \\
5\%   & 44.98 & 0.1895 & 0.0184 & 24.570 \\
15\%  & 48.08 & 0.1969 & 0.0198 & \textit{24.596} \\
25\%  & 49.84 & 0.2015 & 0.0204 & \textbf{24.600} \\
\bottomrule
\end{tabular}
}
\vspace{-5mm}
\end{table}

\section*{Acknowledgments}
This research is funded by Vietnam National University - Ho Chi Minh City (VNU-HCM) under Grant Number B2026-18-17.

\bibliographystyle{IEEEbib}
\bibliography{refs}

\end{document}